\documentclass[11pt]{article}
\usepackage[utf8]{inputenc}
\usepackage[margin=1.2in]{geometry}

\usepackage{amsmath,graphicx}
\usepackage{dsfont}
\usepackage[nolist,nohyperlinks]{acronym}
\usepackage{xspace}
\usepackage{amssymb}
\usepackage{amsfonts}
\usepackage{booktabs}
\usepackage{xcolor}
\usepackage{tikz}
\usepackage[hidelinks]{hyperref}

\newcommand{\eg}{e.g.\xspace}
\newcommand{\ie}{i.e.\xspace}

\newcommand{\Yhat}{\hat{Y}}

\newcommand{\lbd}{$\lambda$\xspace}
\newcommand{\lb}{\lambda}
\newcommand{\lbhat}{\hat{\lambda}}

\newcommand{\Ilb}{{I}_{\lambda}}
\newcommand{\Ilbhat}{{I}_{\hat\lambda}}
\newcommand{\Ulb}{{U}_{\lambda}}
\newcommand{\Ynew}{Y_{\text{new}}}
\newcommand{\Xnew}{X_{\text{new}}}
\newcommand{\Flb}{\mathcal{F}_{\lambda}}
\newcommand{\ntest}{n_{\text{test}}}
\newcommand{\eps}{\varepsilon}
\newcommand{\WW}{W}
\newcommand{\FF}{\mathcal{F}}
\newcommand{\ev}{\mathrm{EV}}

\definecolor{se_col}{HTML}{8ec3f1}
\newcommand{\seCross}[1][0.25]{%
  \begin{tikzpicture}[scale=#1, baseline=0.5ex]
    \fill[se_col] (1,0) rectangle (2,3); 
    \fill[se_col] (0,1) rectangle (3,2); 
    \draw[line width=0.5pt, color=darkgray] (0,0) grid (3,3);
    \fill (1.5,1.5) circle (0.20);
  \end{tikzpicture}%
}

\newcommand{\seSquare}[1][0.25]{%
  \begin{tikzpicture}[scale=#1, baseline=0.5ex]
    \fill[se_col] (0,0) rectangle (3,3); 
    \fill[se_col] (0,0) rectangle (3,3); 
    \draw[line width=0.5pt, color=darkgray] (0,0) grid (3,3);
    \fill (1.5,1.5) circle (0.20);
  \end{tikzpicture}%
}

\definecolor{gtColor}{RGB}{255, 142, 7}
\definecolor{fpColor}{RGB}{216, 27, 96}
\definecolor{tpColor}{RGB}{26, 126, 214}
\definecolor{rejFpColor}{RGB}{0, 77, 64}
\newcommand{\gtPixels}[1][0.5]{%
  \begin{tikzpicture}[scale=#1, baseline=0.4ex]
    \fill[gtColor] (0,0) rectangle (2,2);
    \draw[line width=0.5pt, color=darkgray] (0,0) grid (2,2);
  \end{tikzpicture}%
}
\newcommand{\fpPixels}[1][0.5]{%
  \begin{tikzpicture}[scale=#1, baseline=0.4ex]
    \fill[fill=fpColor] (0,0) rectangle (2,2);
    \draw[line width=0.5pt, color=darkgray] (0,0) grid (2,2);
  \end{tikzpicture}%
}
\newcommand{\tpPixels}[1][0.5]{%
  \begin{tikzpicture}[scale=#1, baseline=0.4ex]
    \fill[fill=tpColor] (0,0) rectangle (2,2);
    \draw[line width=0.5pt, color=darkgray] (0,0) grid (2,2);
  \end{tikzpicture}%
}
\newcommand{\rejFpPixels}[1][0.5]{%
  \begin{tikzpicture}[scale=#1, baseline=0.4ex]
    \fill[fill=rejFpColor] (0,0) rectangle (2,2);
    \draw[line width=0.5pt, color=lightgray] (0,0) grid (2,2);
  \end{tikzpicture}%
}

\author{
    Luca Mossina \qquad Corentin Friedrich
    \vspace{1.25em}
    \\
    IRT Saint Exupéry, Toulouse, France
}

\title{Controlling False Positives in Image Segmentation via Conformal Prediction}
\begin{document}
\date{}
\maketitle
\begin{abstract}
Reliable semantic segmentation is essential for clinical decision making, yet deep models rarely provide explicit statistical guarantees on their errors. 
We introduce a simple post-hoc framework that constructs confidence masks with distribution-free, image-level control of false-positive predictions. 
Given any pretrained segmentation model, we define a nested family of shrunken masks obtained either by increasing the score threshold or by applying morphological erosion. 
A labeled calibration set is used to select a single shrink parameter via conformal prediction, ensuring that, for new images that are exchangeable with the calibration data, 
the proportion of false positives retained in the confidence mask stays below a user-specified tolerance with high probability.
The method is model-agnostic, requires no retraining, and provides finite-sample guarantees regardless of the underlying predictor. 
Experiments on a polyp-segmentation benchmark demonstrate target-level empirical validity.
Our framework enables practical, risk-aware segmentation in settings where over-segmentation can have clinical consequences.
\noindent Code at \url{https://github.com/deel-ai-papers/conseco}.
\end{abstract}

\begin{acronym}
\acro{CP}{{Conformal Prediction}}
\end{acronym}
\newcommand{\cp}{\ac{CP}\xspace}

\section{Introduction}
\label{sec:intro}
Reliable segmentation is a prerequisite for clinical use of deep-learning models, where false positives may trigger unnecessary interventions. 
Existing uncertainty scores and calibration methods provide useful heuristics, but they do not offer finite-sample guarantees on the errors of the produced masks. Given any pretrained segmentation model, our post-hoc method builds inner masks $\Ilb(X)$ by progressively shrinking the predicted mask $\Yhat$ using a single control parameter $\lambda$, through either sigmoid score thresholding or morphological erosion. We calibrate the shrinkage level on a small held-out labeled set.

Our procedure, based on inductive (or ``split'') \cp \cite{Vovk_2005_algorithmic,Papadopoulos_2002_inductive}, guarantees that at a user-chosen confidence level $1 - \alpha$, 
the inner mask contains at most a user-specified fraction $\tau$ of false-positive pixels. 
The validity bound is asserted at the \emph{image level}; 
we refer to the inner mask $\Ilb(X)$ as ``confidence mask'' and the remainder of the prediction is flagged as uncertain, producing the uncertainty region $\Ulb(X) = \Yhat \setminus \Ilb(X)$.

\begin{figure}
    \centering
    \scriptsize
    \includegraphics[width=0.9\linewidth]{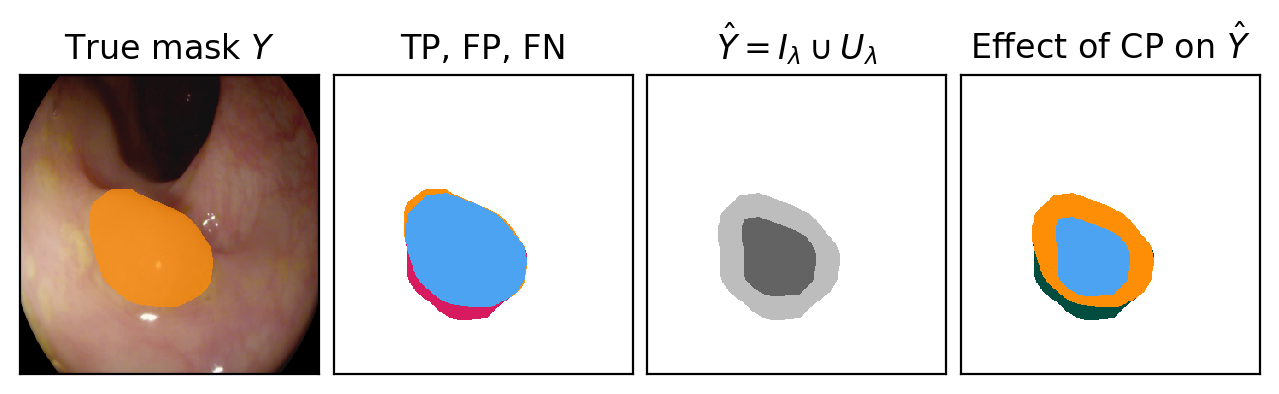}
    \caption{
    Example with erosion inner mask $\Ilb^\eps(X)$ at $\tau=0.01$ and $1-\alpha=0.9$.
    From the left:
    (i) Ground-truth mask $Y$ overlayed on input image $X$;
    (ii) true positives \& false positives in $\Yhat$, and $Y$ pixels missed (false negatives);
    (iii) Inner ``confidence'' mask $\Ilb^\eps(X)$ (dark grey) and uncertainty ``rejection'' mask $\Ulb(X)$ (light grey);
    (iv) $\Yhat$ is shrunk to $\Ilb^\eps(X)$. 
    $\Ulb(X)$ rejects most FPs (\rejFpPixels[0.15]) but also some TPs, \ie g.t. pixels (\gtPixels[0.15]) well-segmented in $\Yhat$.
    ~\\
    \textbf{Colors.}
    \gtPixels[0.15]: true mask $Y$;
    \fpPixels[0.15]: false positives (FP);
    \tpPixels[0.15]: true positives (TP);
    \rejFpPixels[0.15]: rejected FPs. 
    }
    \label{fig:erosion_example}
\end{figure}

\noindent\textbf{Contributions.} 
(i) A black-box, distribution-free method that returns inner prediction sets $\Ilb$ with guaranteed control of accepted false positives at the image level. 
(ii) Two concrete, implementation-ready formulations: score thresholding and morphological erosion. 
(iii) A calibration protocol that exposes two operational choices to the user, confidence $1 - \alpha$ and tolerated accepted false positives $\tau$, with no retraining. 
(iv) Evidence on a biomedical benchmark that the empirical image-level validity of the \textit{accepted false‑positive} proportion (AFP) matches the target confidence.

\section{Related Work}
\label{sec:related_work}
Split \cp \cite{Papadopoulos_2002_inductive} constructs distribution-free prediction sets with finite-sample guarantees of containing the true target at a user-specified confidence level $1-\alpha$.
Conformal Risk Control extends CP to monotone losses, providing guarantees on the expected risk \cite{Angelopoulos_2022_CRC}.
In multilabel prediction, inner and outer sets are constructed to enclose the true label set \cite{Cauchois_2022_multilabel}.

For \cp in semantic segmentation, \cite{Davenport_2024_conformal} have used inner and outer prediction masks targeting coverage of the ground‑truth mask.
Other conformal approaches reduce false negatives by lowering score thresholds \cite{Angelopoulos_2022_CRC,Mossina_2024_varisco,Blot_2025_aa_crc}
or by morphological dilation of the predicted mask 
\cite{Mossina_2025_consema}.
\cp is complementary to the broader literature on uncertainty quantification:
methods such as MC-Dropout \cite{Gal_2016_dropout}, deep ensembles \cite{Lakshminarayanan_2017_ensembles}, or failure prediction \cite{Corbiere_2019_failure_prediction}
provide useful uncertainty maps but do not yield distribution-free guarantees at deployment, which can be achieved with \cp.

\textbf{Our positioning.}
We address binary medical segmentation and control the proportion of accepted false-positive pixels within the predicted region using nested prediction-shrinkers and conformal calibration.
This is complementary to prior false-negative control work and different in scope from multilabel FP-limited set prediction \cite{Fisch_2022_conformal_FP}.
We instantiate the nested sets through standard anti-extensive morphological operators \cite{Serra_1984_math_morpho_v1} and sigmoid-score thresholding.

\section{Methods}
\label{sec:methods}
We aim to control statistically the number of false-positive pixels accepted in predicted masks.
To do so, we define a quantity $\Flb(X,Y)$ (Eq.~\ref{eq:loss_accept_fp}) that is compatible with the requirements of \cp and hence admits rigorous statistical guarantees, notably via the inner sets proposed in Eq.~\ref{eq:inner_set_thresh} and Eq.~\ref{eq:inner_set_erosion}.
We refer to $\Flb$ as the \emph{accepted false-positive proportion (AFP)}: it measures the fraction of false positives that remain in the accepted region relative to the original predicted area $|\Yhat|$.

Let $X$ be an image over a grid $\Omega \subseteq \mathbb{Z}^2$ of $n_H \times n_W$ pixels,
and let $Y \subseteq \Omega$ and $\Yhat \subseteq \Omega$ denote the ground-truth and predicted segmentation masks obtained with a segmentation model, respectively. 

\textbf{Defining inner prediction sets.}
We construct a nested family of \emph{inner prediction sets} 
$\{ \Ilb(X) \}_{\lb \in \Lambda}$
such that:
(i) $\Ilb(X) \subseteq \Yhat$,
(ii) there exists $\lb_0$ such that $I_{\lb_0}(X) = \Yhat$, and
(iii) for any $\lb_1 \le \lb_2$, $I_{\lb_1}(X) \supseteq I_{\lb_2}(X)$.
We interpret the inner prediction sets as \textbf{confidence masks} at a chosen confidence level, defining subregions of the prediction that are ``accepted'' according to the conformal procedure.

We propose two simple ways to shrink a predicted mask so that fewer false-positive pixels are accepted as confident.
We restrict our exposition to two inner set models that are applicable \textit{a posteriori} to most segmentation models, although any nested family of sets (see above) can be used.

First is a set that applies a \textbf{threshold on sigmoid} scores $\hat{\sigma}(X)_{ij}$, the output of a binary segmentation model:
\begin{equation}
\Ilb^{\sigma}(X) 
    := \{\text{pixels } (i,j) \text{ s.t. } \hat \sigma(X)_{ij} \geq \lb \}, 
\label{eq:inner_set_thresh}
\end{equation}
\noindent with $\lb \in [0.5,1]$, where $\lb_0 = 0.5$ is the threshold commonly used in segmentation.

Second, we build a set that works as a dual to the morphological dilation \cite{Serra_1984_math_morpho_v1} used in \cite{Mossina_2025_consema}:
we apply \textbf{morphological erosion} $\eps_B(\cdot)$ to the mask $\Yhat$ as
\begin{equation}    
\Ilb^\eps(X) 
    := \underbrace{(\eps_B \circ \eps_B \circ \dots \circ \eps_B)}_{\lb \text{ iterations}}(\Yhat) 
    = \eps_B^{\lb}(\Yhat).
\label{eq:inner_set_erosion}
\end{equation}
We fix a structuring element, 
\eg $B = \seCross[0.12]$ for 4-connectivity,
and erode $\Yhat$ \lbd times, $\lb \in \mathbb{N}$;
note that for $\lb_0 = 0$, $I_{\lb_0}(X) = \Yhat$.
This erosion model also applies to black-box predictors,
whose internals are not accessible to end users (\eg third-party vendors, embedded in medical equipment, etc.).
This function is suited to segmentation models where the false positives are concentrated at the boundary of the object.
As noted in \cite{Mossina_2025_consema}, \textit{any} morphology-based inner set is applicable, \eg, combining structuring elements or using a discretized ball whose radius is controlled by \lbd. 

\textbf{Our inner sets are nested.}
Our definitions give rise to nested sets, that is, 
for any $\lb_1 \leq \lb_2$, we get $I_{\lb_1} \supseteq I_{\lb_2}$.
For $\Ilb^\sigma(X)$,
as \lbd grows, fewer pixels in $\Yhat$ have scores above this threshold and fewer pixels are included in $\Ilb^\sigma$ (Eq. \ref{eq:inner_set_thresh}), hence its size is non-increasing in \lbd.
For $\Ilb^\eps$,  
since morphological erosion is  anti-extensive (\ie contractive), 
we have $\eps_B^{\lb_1}(\Yhat) \supseteq \eps_B^{\lb_2}(\Yhat)$
for $\lb_1 \leq \lb_2$.

\subsection{Formulation of the False-Positives Control problem}
\label{sec:fp_loss_tolerance}

Inner sets $\Ilb(X)$ aim to ignore false positives in predictions.
However, due to noise in predictive models or annotation errors in segmentation datasets, reaching zero false-positive pixels would require large values of \lbd. 
The obtained inner sets would thus be very small or even empty;
note that the trivial solution $\Ilb = \varnothing$
does not contain any FPs and it is always valid.
To avoid this trivial solution, we allow a small fraction of FPs,
which is controlled by a user-defined tolerance parameter $\tau \in [0,1]$.

Let 
$\WW(X,Y) = \Yhat \cap (\Omega \setminus Y)$ 
denote the set of \textbf{false-positive} (FP) pixels in $\Yhat$.
This set does not depend on the inner mask $\Ilb$.
Since we cannot control the size of $\WW(X,Y)$, which depends on the fixed predictor,
we instead construct inner masks that are the largest subsets of $\Yhat$ containing few false positives,
according to the chosen definition of inner mask (Eqs.~\ref{eq:inner_set_thresh}--\ref{eq:inner_set_erosion}).
We refer to $\Ilb(X)$ as the \emph{confidence mask} and to its complement
$\Ulb(X) = \Yhat \setminus \Ilb(X)$ as the \emph{uncertain region}.

We want to control the following quantity, representing the \textbf{accepted false-positive proportion} (AFP) within the predicted mask $\Yhat$:
\begin{equation}
    \Flb(X,Y)
    =
    \frac{|\Ilb(X)\cap\WW(X,Y)|}{|\Yhat|}.
    \label{eq:loss_accept_fp}
\end{equation}
Here $|\cdot|$ denotes set cardinality.
If $|\Yhat|=0$, we set $\Ilb=\varnothing$ and $\Flb=0$.
Since $\Ilb$ is nested and the denominator $|\Yhat|$ is $\lb$-invariant,
$\Flb$ is non-increasing in $\lb$.
Dividing by $|\Ilb|$ would break this monotonicity and is therefore avoided.

\textbf{Interpretation.}
$\Flb$ quantifies the proportion of false positives that remain ``accepted''
within the confidence mask at a level $\lb$.
For $\lb=\lb_0$ (no shrinkage), $\mathcal F_{\lb_0}=|\WW(X,Y)|/|\Yhat|$
corresponds to the original false-positive fraction of the prediction.
Increasing $\lb$ enforces stricter acceptance and can only decrease $\Flb$.
In practice, $\Flb$ expresses the fraction of spurious detections that remain unfiltered after applying the confidence threshold;
in some cases it is possible to have $\Flb(X,Y) \leq \tau$, in which case no shrinkage would be needed.

\subsection{Conformal Prediction}
\label{sec:cp-methods}
Our method builds on the standard inductive \cp framework \cite{Papadopoulos_2002_inductive}.
We adapt the inner sets from the segmentation approach of \cite{Davenport_2024_conformal}, 
and use morphological erosion as the counterpart of the dilation-based outer sets in \cite{Mossina_2025_consema}.
Importantly, 
\cp\ provides \textit{marginal frequentist} guarantees on the \textit{mask-level procedure}, rather than on individual pixels: 
if the calibration and testing process were repeated many times, 
the empirical validity condition in Eq.~\eqref{eq:cp_guarantee} would be satisfied in at least $100(1-\alpha)\%$ of cases on average.
This states that, for exchangeable data, the accepted false-positive proportion (AFP) of the inner mask satisfies \(\mathcal F_{\lbhat}\le \tau\) with probability at least \(1-\alpha\) at the image level.
Once the user has fixed a tolerance value $\tau \in [0,1]$, 
the nonconformity score $r_\iota = r(X_\iota, Y_\iota)$ 
for a calibration pair $(X_\iota, Y_\iota)$ 
is computed as (details in Sec.~\ref{sec:cp_algo})
\begin{equation}
    r_\iota = 
    \inf \left\{\,\lb \,:\, \Flb(X_\iota, Y_\iota) \leq \tau \right\},
    \label{eq:nonconformity_score}
\end{equation}
and the conformalizing quantile is the
\begin{equation}
    \hat{\lb} = 
    \lceil (n+1)(1-\alpha) \rceil\text{-th smallest value in } 
    (r_\iota)_{\iota=1}^{n}.
    \label{eq:conformal_quantile}
\end{equation}
Assuming that calibration data and test point $(\Xnew, \Ynew)$ form an exchangeable (or i.i.d.) sequence,
we obtain the distribution-free, marginal guarantee:
\begin{equation}
    \mathbb{P}
    \left(
        \FF_{\lbhat}(\Xnew, \Ynew) \leq \tau
    \right)
    \geq 1 - \alpha.
    \label{eq:cp_guarantee}
\end{equation}

\textbf{Proof}. 
Because inner masks $\Ilb(X)$ are nested with respect to \lbd,  $\Flb$ is non-increasing in \lbd.
Define the binary loss 
$\ell(X,Y,\lb) = \mathds{1}\!\left\{
        \Flb(X,Y) > \tau
    \right\}
$, 
which is monotone non-increasing in \lbd.
Applying Conformalized Risk Control \cite{Angelopoulos_2022_CRC} with this loss, whose CRC selection rule coincides with the conformal quantile (Sec.~2.3 in \cite{Angelopoulos_2022_CRC}), yields Eq.~\eqref{eq:cp_guarantee}.

\textbf{Interpretation.}
This image-level guarantee states that, 
with probability (\ie confidence level) at least $1 - \alpha$, 
the region of the inner mask $\Ilbhat(\Xnew)$ 
will not ``accept'' more than a fraction $\tau$ (\eg, $5\%$) false positives in the original prediction $\hat Y_{\text{new}}$.
This can also be seen as a mechanism to produce confidence regions that mitigate the operational hazard of predicting a treatment (\ie a ``positive'' pixel) where not needed.
Clinically, AFP bounds the fraction of unnecessary positive pixels \emph{within the region we agree to act upon}, while the rejected pixels are marked as uncertain and not acted on.

\textbf{Scope.}
Our guarantee depends only on exchangeability of the data and on the nestedness of the inner sets ${\Ilb(X)}_{\lb \in \Lambda}$.
It does not depend on the particular way $\Ilb$ is produced.
Any scoring function or post-hoc shrinker that induces a nested family can be plugged in; this choice affects only \emph{utility} (e.g., average contraction, inner-margin size, ATP/CR), not validity.
The convention of a fixed denominator $|\widehat Y|$ and the handling of empty masks ($\mathcal F_\lambda=0$) likewise do not change the guarantee;
they only influence how conservative the resulting inner masks are.
This opens the method to empirical exploration of shrinkers and scores from any uncertainty model.

\subsection{Conformalization procedure}
\label{sec:cp_algo}

One can use any fixed segmentation model, either pretrained or trained separately on dedicated data.

\begin{enumerate}
\item \textbf{Collect calibration data.}
Gather a labeled calibration set $\{(X_\iota, Y_\iota)\}_{\iota=1}^{n}$, distinct from both training and test data.
Following standard \cp assumptions, these samples are required to be exchangeable (or i.i.d.) with the test cases.
\item \textbf{Define the family of inner sets.}
Select either: (i) the threshold model $\Ilb^{\sigma}$ in Eq.~\eqref{eq:inner_set_thresh} with $\Lambda = [0.5,1]$ 
or 
(ii) the morphological model $\Ilb^{\eps}$ in Eq.~\eqref{eq:inner_set_erosion} with $\Lambda = \mathbb{N}$ and a structuring element, \eg $B = \seCross[0.12]$ or \seSquare[0.12].
\item \textbf{Compute calibration scores.}
For each calibration pair $(X_\iota, Y_\iota)$, compute $r_\iota = r(X_\iota, Y_\iota)$ as in Eq.~\eqref{eq:nonconformity_score}.
\begin{itemize}
    \item [(i)] For $\Ilb^{\sigma}$, evaluate $\Flb$ at breakpoints 
    $\lb \in \{\hat\sigma(X_\iota)_{ij} : (i,j)\in \Yhat_\iota\} \cup \{1\}$ 
    and keep the smallest \lbd s.t. 
    $\Flb \le \tau$.
    \item [(ii)] For $\Ilb^{\eps}$, iterate $\eps_B$ until $\Flb \le \tau$ or the mask becomes empty, in which case $\Flb = 0$.
\end{itemize}
\item \textbf{Estimate the conformal threshold.}
Let $k = \lceil (n+1)(1-\alpha) \rceil$,
and set $\lbhat$ to the $k$-th smallest value among the calibration scores $\{r_\iota\}_{\iota=1}^{n}$, as in Eq.~\eqref{eq:conformal_quantile}.
\item \textbf{Predict on the test set.}
For a new input $X_{\mathrm{new}}$, output the inner mask $I_{\lbhat}(X_{\mathrm{new}})$. 
\end{enumerate}

\section{Experiments}
\label{sec:expe}

Following prior work on distribution-free methods for segmentation
\cite{Bates_2021_RCPS,Angelopoulos_2022_CRC,Blot_2025_aa_crc},
we evaluate on the \textsc{Polyps} dataset collection
\cite{Silva_2014_ETIS_dataset,
Bernal_2015_CVC_Clinic_Dataset,
Tajbakhsh_2015_CVC_ColonDB,
Vazquez_2017_EndoScene,
Pogorelov_2017_kvasir_data}
using pretrained PraNet \cite{Fan_2020_pranet}.\footnote{
We use the precomputed predictions as distributed by the authors of \cite{Bates_2021_RCPS,Angelopoulos_2022_CRC} for their comprehensive introduction to \cp \cite{Angelopoulos_2021_Gentle} at \textit{github.com/aangelopoulos/conformal-prediction}, 
}
We compare three procedures: \emph{Baseline} (uncalibrated sigmoid threshold at $0.5$, i.e., $I_{\lb_0}=\Yhat$), \emph{Threshold} (CP by increasing the score cutoff), and \emph{Erosion} (CP by morphological erosion). 
Confidence level is $1-\alpha = 0.90$ and tolerance $\tau\in\{0.1,0.01,0.001\}$. 
For erosion we use a cross structuring element $B=\seCross[0.12]$ . 
Each run randomly permutes the test split and assigns half to calibration and half to evaluation, yielding $250$ calibration images. 
We report mean $\pm$ standard deviation over $10$ seeds.

\textbf{Metrics.}
We quantify validity and utility with three image-level quantities. 
(i) \textit{Empirical validity} (EV): fraction of test images whose accepted false-positive proportion does not exceed $\tau$, 
that targets the nominal frequency $1-\alpha$,
\begin{equation}
\ev = \frac{1}{\ntest}\sum_{i=1}^{\ntest}\mathds{1}\!\left\{\FF_{\lbhat}(X_i, Y_i)\le\tau\right\}.
\end{equation}
\noindent
(ii) \textit{Contraction ratio} (CR): retained area after shrinkage,
\begin{equation}
\mathrm{CR}=\frac{1}{\ntest}\sum_{i=1}^{\ntest}\frac{|I_{\lbhat}(X_i)|}{|\Yhat_i|}.
\end{equation}
CR $= 1$ for the Baseline and decreases as masks shrink, measuring the utility loss incurred by enforcing statistical validity.

\noindent
(iii) \textit{Accepted true-positive fraction} (ATP):
\begin{equation}
\mathrm{ATP}=\frac{1}{\ntest}\sum_{i=1}^{\ntest}\frac{|I_{\lbhat}(X_i)\cap Y_i|}{|\Yhat_i|}.
\end{equation}
For the Baseline, $\mathrm{ATP}=\frac{|TP|}{|\Yhat|}$ equals image-level precision, \ie the fraction of TPs in the prediction mask.

\subsection{Results}
\label{sec:results}
\begin{figure}[t!]
    \centering
    \scriptsize
    \includegraphics[width=0.9\linewidth]{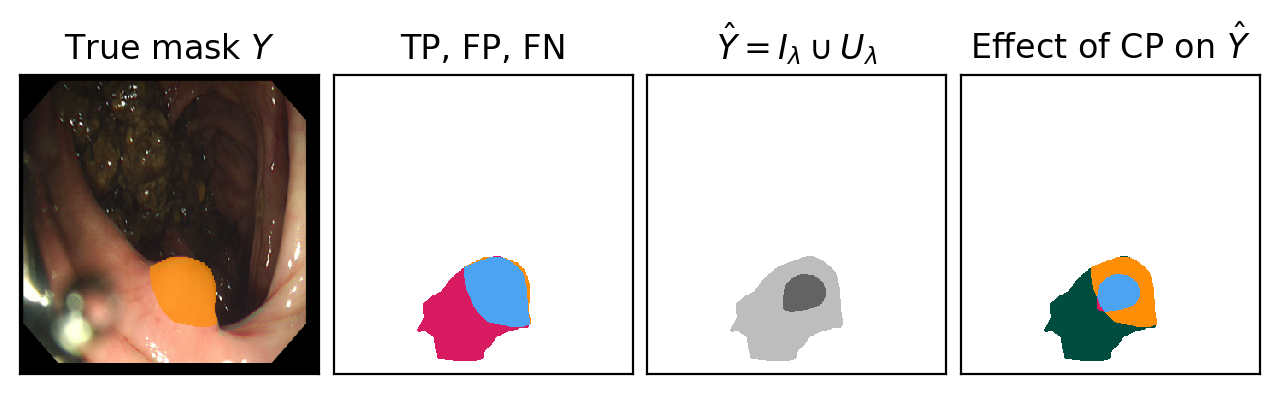}
    \includegraphics[width=0.9\linewidth]{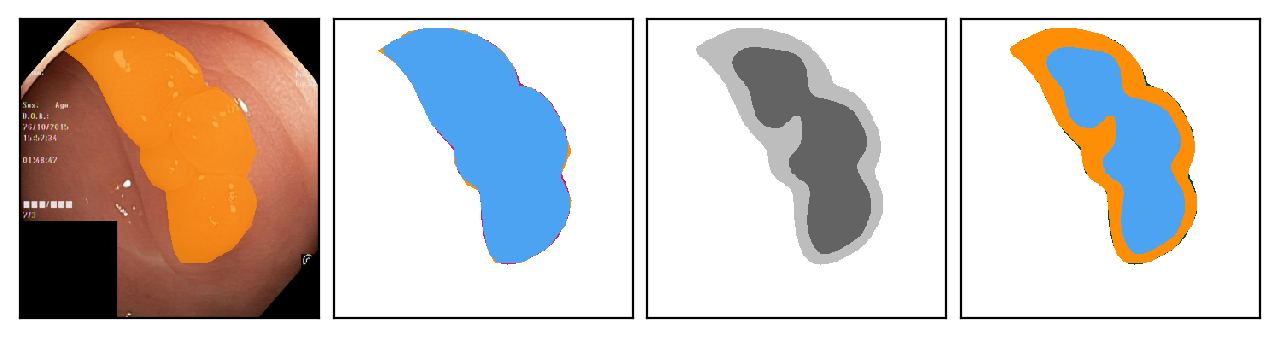}
    \includegraphics[width=0.9\linewidth]{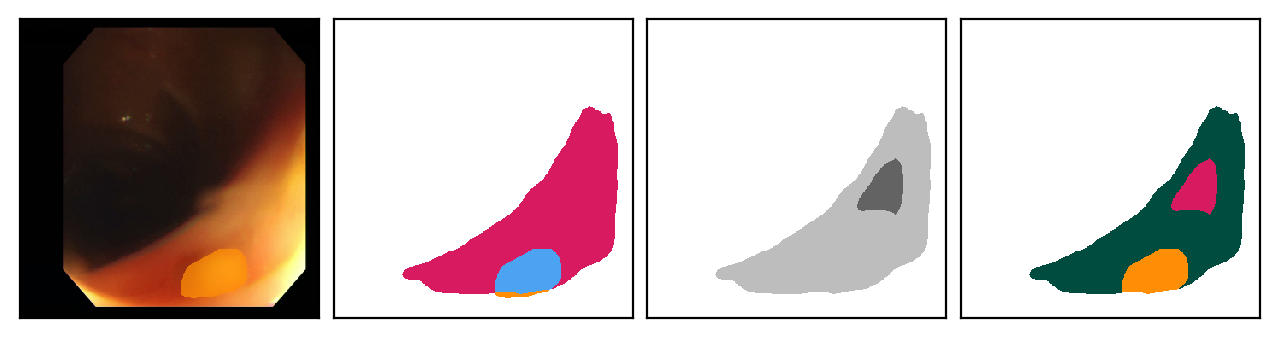}
    \caption{Examples with thresholding inner mask $\Ilb^\sigma(X),$ at $\tau=0.01$ and $1-\alpha=0.9$. 
    \textbf{Top}. Large FP removal with moderate TP loss. 
    \textbf{Middle}. When FPs are already negligible, shrinkage removes TPs. 
    \textbf{Bottom}. Failure case: residual inner mask is FP-only. 
    ~\\
    \textbf{Colors}. 
    \gtPixels[0.15]: true mask $Y$; 
    \fpPixels[0.15]: false positives; 
    \tpPixels[0.15]: true positives;
    \rejFpPixels[0.15]: rejected false positives.}
    \label{fig:thresh_example}
\end{figure}

\begin{table}[ht]
\centering
\small
\setlength{\tabcolsep}{2pt}
\begin{tabular}{llccc}
 & \multicolumn{1}{c}{Method} & $\ev$ & CR ($\uparrow$) & ATP ($\uparrow$) \\
\midrule
$\tau=0.1$ & Baseline & 0.789 ± {\scriptsize 0.015} & --- & 0.873 ± {\scriptsize 0.014} \\
           & Threshold & {0.931} ± {\scriptsize 0.049} & 0.701 ± {\scriptsize 0.129} & 0.676 ± {\scriptsize 0.111} \\
           & Erosion & {0.897} ± {\scriptsize 0.030} & 0.578 ± {\scriptsize 0.155} & 0.527 ± {\scriptsize 0.132}  \vspace{0.5em} \\
$\tau=0.01$ & Baseline & 0.165 ± {\scriptsize 0.013} & --- & 0.873 ± {\scriptsize 0.014} \\
 & Threshold & 0.926 ± {\scriptsize 0.037} & 0.532 ± {\scriptsize 0.087} & 0.525 ± {\scriptsize 0.083} \\
 & Erosion & 0.902 ± {\scriptsize 0.034} & 0.274 ± {\scriptsize 0.085} & 0.252 ± {\scriptsize 0.075}  \vspace{0.5em} \\
$\tau=0.001$ & Baseline & 0.005 ± {\scriptsize 0.002} & --- & 0.873 ± {\scriptsize 0.014} \\
 & Threshold & 0.914 ± {\scriptsize 0.029} & 0.473 ± {\scriptsize 0.067} & 0.469 ± {\scriptsize 0.065} \\
 & Erosion & 0.911 ± {\scriptsize 0.033} & 0.112 ± {\scriptsize 0.052} & 0.103 ± {\scriptsize 0.045}  \vspace{0.5em} \\
\end{tabular}
\caption{Results for confidence level $1 - \alpha = 0.9$.}
\label{tab:polyps_a10}
\end{table}

Quantitative results at $1-\alpha = 0.9$ are reported in Tab.~\ref{tab:polyps_a10}. 
Conformalized procedures attain empirical validity (EV) close to the nominal target across all $\tau$, whereas the Baseline fails for small $\tau$ and only approaches validity in the more permissive setting $\tau = 0.1$, confirming the need for calibration.
Utility decreases smoothly as $\tau$ tightens. 
At fixed $\tau$, the Threshold variant retains higher CR and ATP than Erosion, reflecting a more moderate shrinkage path on typical polyp masks, while Erosion removes a larger fraction of peripheral predictions. 
This is expected, 
as sigmoid scores are often more informative than the binary mask; when available, they should be evaluated alongside erosion.
The validity-utility trade-off behaves monotonically and remains stable across random seeds. 
Finally, since Conformal Prediction guarantees statistical validity for \textit{any} segmentation model, utility metrics can also serve to compare or select among different predictors, \eg when provided by a third-party.

\section{Conclusion and Perspectives}
\label{sec:conclusion}
We introduced a post-hoc conformal procedure for binary segmentation that calibrates a single shrink parameter to control, with finite-sample guarantees, the image-level proportion of accepted false positives. 
The method is model-agnostic, requires no retraining, and operates with either score-thresholding or morphological-erosion shrinkers. 
Experiments on polyp segmentation demonstrate target-level empirical validity and a smooth validity–utility trade-off governed by $\tau$. 
Limitations include the mask-level (marginal) nature of the guarantees and potential true-positive loss when predictions are already precise. 
Future work includes extending the approach to multi-class segmentation and building size- and instance-adaptive inner masks.

\subsubsection*{Acknowledgments}
This work was carried out within the DEEL project,\footnote{https://www.deel.ai} which is part of IRT Saint Exupéry and the ANITI AI cluster. The authors acknowledge the financial support from DEEL's Industrial and Academic Members and the France 2030 program – Grant agreements n°ANR-10-AIRT-01 and n°ANR-23-IACL-0002.

\bibliographystyle{acm}
\bibliography{biblio}

\end{document}